\documentclass[12pt]{article}
\newcommand{\blind}{0}
\usepackage[nopatch=footnote]{microtype}
\usepackage{graphicx}
\usepackage{subcaption}
\usepackage{booktabs} 

\usepackage{hyperref}

\usepackage{amsmath}
\usepackage{amssymb}
\usepackage{mathtools}
\usepackage{amsthm}

\usepackage[capitalize,noabbrev]{cleveref}

\theoremstyle{plain}
\newtheorem{theorem}{Theorem}
\newtheorem{proposition}[theorem]{Proposition}

\theoremstyle{definition}

\newtheorem{assumption}[theorem]{Assumption}
\theoremstyle{remark}
\newtheorem{remark}[theorem]{Remark}

\usepackage[textsize=tiny]{todonotes}

\usepackage{enumerate}
\usepackage{algorithm}
\usepackage{algpseudocode}
\usepackage{bbm}
\usepackage{multirow}

\newcommand{\cR}{\mathcal{R}}
\newcommand{\bG}{\mathbb{G}}

\newcommand{\bH}{\mathbb{H}}
\newcommand{\bP}{\mathbb{P}}
\newcommand{\cG}{\mathcal{G}}

\newcommand{\cK}{\mathcal{K}}
\newcommand{\cJ}{\mathcal{J}}

\newcommand{\cH}{\mathcal{H}}
\newcommand{\R}{\mathbb{R}}
\newcommand{\wt}{\widetilde}
\newcommand{\wh}{\widehat}

\newcommand{\id}{\mathbbm{1}}
\newcommand{\E}{\mathbb{E}}

\DeclareMathOperator*{\var}{Var}

\newcommand{\li}{\langle}
\newcommand{\ri}{\rangle}

\usepackage{natbib} 

\usepackage[nodisplayskipstretch]{setspace}

\begin{document}

        \def\spacingset#1{\renewcommand{\baselinestretch}%
		{#1}\small\normalsize} \spacingset{1}
	
        \if0\blind
        {
            \title{\bf Testing Conditional Mean Independence Using Generative Neural Networks}
            \author{Yi Zhang\thanks{Equal contribution}, \hspace{.1cm} Linjun Huang\footnotemark[1], \hspace{.1cm} Yun Yang and Xiaofeng Shao
                \thanks{Yi Zhang and Linjun Huang are Ph.D. students at Department of Statistics, University of Illinois at Urbana-Champaign, Champaign, USA. Yun Yang is Associate Professor at Department of Mathematics, University of Maryland, College Park MD, USA. Xiaofeng Shao is Professor at Department of Statistics and Data Science, Washington University in St Louis, USA. Emails: {\tt yiz19@illinois.edu}, {\tt linjunh2@illinois.edu}, {\tt yy84@umd.edu} and {\tt shaox@wustl.edu}. }\\
            }
            \maketitle
        } \fi
	
	\if1\blind
	{
		\bigskip
		\bigskip
		\bigskip
		\begin{center}
			{\LARGE\bf Title}
		\end{center}
		\medskip
	} \fi
	
	\bigskip

		\begin{abstract}
			
			Conditional mean independence (CMI) testing is crucial for statistical tasks including model determination and variable importance evaluation. In this work, we introduce a novel population CMI measure and a bootstrap-based testing procedure that utilizes deep generative neural networks to estimate the conditional mean functions involved in the population measure. The test statistic is thoughtfully constructed to ensure that even slowly decaying nonparametric estimation errors do not affect the asymptotic accuracy of the test. Our approach demonstrates strong empirical performance in scenarios with high-dimensional covariates and response variable, can handle multivariate responses, and maintains nontrivial power against local alternatives outside an $n^{-1/2}$ neighborhood of the null hypothesis. We also use numerical simulations and real-world imaging data applications to highlight the efficacy and versatility of our testing procedure.

		\end{abstract}

        	\noindent%
	\vfill
		
		\section{Introduction}
		
	Conditional mean independence (CMI) testing is a fundamental tool for model simplification and assessing variable importance, and plays a crucial role in statistics and machine learning. In traditional statistical applications, such as nonparametric regression, CMI testing identifies subsets or functions of covariates that meaningfully predict the response variable. This is essential for improving model efficiency, accuracy, and interpretability by avoiding redundant variables. In machine learning, CMI testing has broad applications in areas like interpretable machine learning \citep{murdoch2019interpretable}, representation learning \citep{bengio2013representation, huang2024deep} and transfer learning \citep{maqsood2019transfer, zhuang2020comprehensive}.
	
    In this paper, we address the problem of CMI testing for multivariate response variables and covariates. Specifically, for random vectors $X \in \R^{d_X}$, $Y \in \R^{d_Y}$, and $Z \in \R^{d_Z}$, 
    we test the null hypothesis $H_0$ that $Y$ is conditionally mean independent of $X$ given $Z$, i.e., $\E[Y|X=x,Z=z] = \E[Y|Z=z]$, a.e. $(x,z) \in \R^{d_X+d_Z}$.
For example, consider predicting age $Y$ from a facial image. To test whether $Y$ can be predicted using images with a specific facial region (potentially containing sensitive information) covered in black, $X$ can represent the covered region and $Z$ the remaining facial image. Similarly, to test whether $Y$ can be predicted using a low-resolution version or extracted features of a facial image, $X$ can represent the original image and $Z$ a low-dimensional feature vector derived from $X$ using standard extraction methods (e.g., Autoencoder, Average pooling, or PCA \citep{berahmand2024autoencoders}). Under $H_0$, removing $X$ from the predictive model $f(X,Z)$, where $f: \R^{d_X+d_Z} \to \R^{d_Y}$ denotes the nonparametric regression function, does not reduce prediction accuracy.

For machine learning tasks, identifying which (functions of) covariates contribute to predicting the response is particularly important, as deep neural networks (DNNs) often process high-dimensional data, such as images and text, that may include irrelevant features \citep{li2017example,van2010discriminative}. Performing dimensionality reduction before DNN training provides several advantages: it enhances interpretability, improves prediction accuracy, and reduces computational costs \citep{cai2018feature}. In evaluating DNN training, covariate significance is typically assessed by comparing performance metrics, such as mean squared error or classification accuracy, for DNNs trained with and without specific covariates \citep{dai2022significance, williamson2023general}.
As discussed in Section 2.3 of \citet{williamson2023general}, many common performance metrics, such as the coefficient of determination ($R^2$) for regression and empirical risk with cross-entropy or 0-1 loss for binary classification, are functionals of the conditional mean functions of $Y$, which means $H_0$ will imply that the covariates are not significant. Thus, CMI testing provides a valuable tool for assessing covariate importance across a variety of machine learning applications.

		\subsection{Related literature}

To the best of our knowledge, most existing CMI tests focus on univariate $Y$ and face one or more of the following three major issues: (1) finite-sample performance deteriorates when some or all of $(X, Y, Z)$ are high-dimensional; (2) the tests lack theoretical size guarantees in general nonparametric settings; and (3) they exhibit weak power in detecting local alternatives. For a recent survey, see \citet{lundborg2023modern}. 
We discuss how these challenges arise and how existing CMI tests have partially addressed them.
		
		\noindent {\bf Performance deterioration in high dimensional setting.} 
   This issue primarily arises from the estimation of the conditional mean functions $r(z):=\E[Y|Z=z]$ and $m(x,z):=\E[Y|X=x,Z=z]$. Early CMI tests, such as those in \citet{fan1996consistent, delgado2001significance, zhu2012dim, lavergne2000nonparametric, ait2001goodness}, relied on kernel smoothing methods for these estimations.
	Consequently, these CMI tests suffer from the curse of dimensionality: their performance declines significantly as the dimensions $d_Z$, $d_X+d_Z$, and/or $d_Y$ increase \citep[Section 1]{zhou2022deep}. For instance, Figure 1 in \citet{zhu2012dim} shows that the empirical power of tests proposed by \citet{fan1996consistent} and \citet{delgado2001significance} decreases sharply with increasing $d_Z$ and $d_X$, exhibiting trivial power under a sample size of $n=200$ with moderate dimensionality ($d_Z=4$ and $d_X \geq 8$). To address this, recent CMI tests utilize machine learning tools, such as deep neural networks (DNNs) and the kernel trick, to estimate conditional mean functions. These tools are effective in approximating complex, high-dimensional functions with underlying low-dimensional structures. For example, tests in \citet{williamson2021nonparametric, dai2022significance, lundborg2022projected, cai2022model, williamson2023general, cai2024partial} leverage DNNs for conditional mean estimation, achieving better performance in high-dimensional settings.
		
		\noindent {\bf Theoretical size guarantee.} The main challenge in establishing a theoretical size guarantee stems from two key issues. First, most of the existing CMI tests (except the one proposed by \citet{delgado2001significance}) rely on the sample estimation of the population CMI measure $\Gamma := \E\big[ (r(Z) - m(X, Z))^2 w(X, Z) \big]$ or its equivalent forms, where $w$ is a positive weight function. A common plug-in estimator of $\Gamma$ is given by $\wh \Gamma (\wh r, \wh m) = n^{-1} \sum_{i=1}^n (\wh r(Z_i) - \wh m(X_i, Z_i))^2 w(X_i, Z_i)$, where $\wh r$ and $\wh m$ are estimators of the conditional mean functions. For a population CMI measure to be valid, it must uniquely characterize CMI, meaning that the measure equals zero if and only if $H_0$ holds. While $\Gamma$ satisfies this requirement, its estimator $\wh \Gamma (\wh r, \wh m)$ suffers from a degeneracy problem: under $H_0$, $\wh \Gamma (\wh r, \wh m)$ converges to zero at a rate faster than the $n^{-1/2}$ rate at which $\wh \Gamma(r, m) - \Gamma$ converges to a non-degenerate limiting distribution under the alternative \citep[Section 1]{fan1996consistent}.
		
       Second, the nonparametric estimation errors for $r(z)$ and $m(x, z)$ typically decay slower than the $n^{-1/2}$ parametric rate, and the convergence rate of $\wh \Gamma (\wh r, \wh m)$ under $H_0$ depends heavily on how quickly these errors decay. For CMI tests that use kernel smoothing to estimate the conditional mean functions, the estimation error has an explicit form, allowing it to be addressed directly, along with the degeneracy issue, when deriving asymptotic results under $H_0$. As a result, all the aforementioned kernel smoothing-based CMI tests have theoretical size guarantees.
       In contrast, due to the black-box nature of DNNs, the estimation errors for $r(z)$ and $m(x, z)$ cannot be explicitly decomposed or handled in the same way as kernel smoothing estimators. Consequently, addressing the degeneracy issue requires additional debiasing procedures to mitigate the impact of these errors and achieve accurate size control. For example, \citet{williamson2021nonparametric} and \citet{lundborg2022projected} constructed statistics based on alternative forms of $\Gamma$ to reduce bias. However, the degeneracy issue persists: as shown in Theorem 1 of \citet{williamson2021nonparametric}, their estimator is $\sqrt{n}$-consistent when $\Gamma > 0$, but no asymptotic results are provided under $H_0$ (i.e., $\Gamma = 0$). Similarly, the test in \citet{lundborg2022projected}, which builds on the conditional independence test from \citet{Shah2020hard}, relies on strong assumptions to bypass the degeneracy issue, as outlined in Section 2.1 and Part (a) of Theorem 4 in \citet{lundborg2022projected}.
       To address the degeneracy issue, \citet{dai2022significance} introduced additional noise of order $O_p(n^{-1/2})$ to the estimator of $\Gamma$, while \citet{williamson2023general} utilized sample splitting to estimate separate components of an equivalent form of $\Gamma$ on different subsamples. However, both approaches are ad hoc (Section 6.2 of \citet{verdinelli2024feature}) and, as discussed in Appendix S3 of \citet{lundborg2022projected}, suffer from significant power loss under the alternative hypothesis.
		
		\noindent {\bf Weak power against local alternatives.} On one hand, the CMI tests proposed in \citet{fan1996consistent, zhu2012dim, lavergne2000nonparametric, ait2001goodness, williamson2021nonparametric, dai2022significance, williamson2023general} fail to detect local alternatives with signal strength $\Delta_n := \sqrt{\E[(r(Z) - m(X, Z))^2]}$ of order $n^{-1/2}$, primarily due to their reliance on the population measure $\Gamma$. Specifically, the tests in \citet{fan1996consistent, zhu2012dim, lavergne2000nonparametric, ait2001goodness} take the form $nh^{s/2}\wh \Gamma$, where $h \to 0$ is a bandwidth parameter used in kernel smoothing, and $s = d_Z$ or $d_X + d_Z$. Consequently, these tests cannot detect local alternatives converging to the null faster than $n^{-1/2}h^{-s/4}$. The tests in \citet{williamson2021nonparametric, dai2022significance, williamson2023general} use the population CMI measure $\Gamma_0 = \Gamma_1 - \Gamma_2$, where $\Gamma_1 = \E[(Y - r(Z))^2]$ and $\Gamma_2 = \E[(Y - m(X, Z))^2]$, which is equivalent to $\Gamma$. Since the quadratic terms $\Gamma_1$ and $\Gamma_2$ can only be estimated at the $n^{-1/2}$ rate, these tests are limited to detecting local alternatives with $\Delta_n$ of order $n^{-1/4}$. On the other hand, the test proposed in \citet{cai2024partial} employs sample splitting and requires the size of the training subsample used to estimate the conditional mean functions to be substantially larger than that of the testing subsample. As a result, their test is limited to detecting local alternatives with $\Delta_n$ converging to zero slower than $n^{-1/2}$, which can still result in significant power loss in practice.

		\subsection{Our contributions}
		
		\begin{table}[t]
			\caption{Summary of existing CMI tests. \textsf{High-Dim}: whether the test has good empirical performance when some or all the dimensions of $X$, $Y$ and $Z$ are large; \textsf{Size Guarantee}: whether the test has theoretical results under $H_0$ that guarantee accurate asymptotic size control in general nonparametric setting; \textsf{Local Alt}: whether the test can detect local alternatives with signal strength $\Delta_n$ converging to zero at the parametric rate $n^{-1/2}$.    }\label{tab_11}
			\centering
			\scriptsize
			\setlength{\tabcolsep}{4.5pt}
			\begin{tabular}{c|cccc}
				\toprule
				\midrule
				Tests & \textsf{High-Dim} & \textsf{Size Guarantee} & \textsf{Local Alt} \\\hline
				\citet{fan1996consistent}    & No& Yes & No \\\hline
				\citet{delgado2001significance}  & No& Yes & Yes \\\hline
				\citet{zhu2012dim}  &No & Yes & No\\\hline
				\citet{lavergne2000nonparametric}  &No &Yes & No\\\hline
				\citet{ait2001goodness}  &No & Yes& No\\\hline
				\citet{williamson2021nonparametric} &Yes & No & No\\\hline
				\citet{ dai2022significance} &Yes &Yes & No\\\hline
				\citet{lundborg2022projected} &Yes & No& Yes\\\hline
				\citet{williamson2023general} &Yes & Yes & No\\\hline
				\citet{cai2024partial} &Yes & Yes & No\\\hline
				\textcolor{blue}{\bf{Our method}} &\textcolor{blue}{\bf{Yes}} & \textcolor{blue}{\bf{Yes}} & \textcolor{blue}{\bf{Yes}}\\\hline			
			\end{tabular}
			\normalsize
		\end{table}

		Table \ref{tab_11} summarizes limitations of existing CMI tests. To overcome these challenges, we propose a novel CMI testing procedure with the following advantages:
		\setlength{\itemsep}{-1ex}
		\setlength{\parskip}{0ex}
		\setlength{\topsep}{0ex}
		\setlength{\partopsep}{1ex}
		\setlength{\parsep}{1ex}
		\begin{enumerate} \vspace{-0.5em}
			\item The test demonstrates strong empirical performance even when some or all dimensions of $X$, $Y$, and $Z$ are high. Notably, it is well-suited for scenarios where imaging data serve as covariates, responses, or both.
			 \vspace{-0.5em}
			\item The test achieves precise asymptotic size control under $H_0$. \vspace{-0.5em}
			\item  The test exhibits nontrivial power against local alternatives outside an $n^{-1/2}$-neighborhood of $H_0$. \vspace{-0.5em}
		\end{enumerate}

To achieve these features, we propose a new population CMI measure closely related to the conditional independence measure introduced in \citet{daudin1980partial}. Additionally, we develop a sample version of this population measure in a multiplicative form, which is key to mitigating the impact of estimation errors in nonparametric nuisance parameters (i.e., the conditional mean functions).
Our test not only requires estimating $r(z)$ but also the conditional mean embedding \citep[CME,][Definition 3]{songle2009} of $X$ given $Z$ into a reproducing kernel Hilbert space (RKHS) on the space of $X$. Instead of directly estimating the CME using DNNs, we train a generative neural network (GNN) to sample from the (approximated) conditional distribution of $X$ given $Z$. The CME is then estimated using the Monte Carlo method with samples generated from the trained GNN.
		
		\subsection{Organization and notations}
The paper is organized as follows: Section \ref{sec2_method} introduces the proposed population CMI measure, the test statistic, and the bootstrap calibration procedure, along with its asymptotic properties and consistency results. Section \ref{sec3_simu} evaluates the test using finite sample simulations, comparing it with other methods. Section \ref{sec5_real} presents two real data examples. Section \ref{sec6conclu} concludes with final remarks. All proofs and additional details are deferred to the Appendix.

The following notations will be used throughout the paper. For any positive integer $d$ and random vectors $(X^1, X^2, \dots, X^d, Z)$ defined on the same probability space, $\bP_{X^1X^2\cdots X^d}$ and $\bP_{X^1X^2\cdots X^d|Z}$ denote the joint distribution of $X^1, X^2, \dots, X^d$ and its conditional distribution given $Z$, respectively. Let $\E_Z$ represent the expectation with respect to $\bP_Z$, and let $\bP_m$ denote the Lebesgue measure on $\R^d$. For a positive integer $n$, define $[n] = \{1, 2, \dots, n\}$. For a probability measure $\mu(\cdot)$ on $\R^d$ and $p \geq 1$, let $L_p(\R^d, \mu) = \{f : \R^d \to \R : \int |f(x)|^p \, d\mu(x) < \infty\}$. For any Hilbert spaces $A$ and $B$, let $A \otimes B$ denote their tensor product, and use $\li\,\cdot\,,\,\cdot\,\ri_{A}$ and $\|\cdot\|_A$ to denote the inner product and induced norm on $A$, respectively. For random vectors $a, b \in \R^d$, the Gaussian kernel is defined as $\cK(a, b) = \exp(-\|a-b\|_2^2/(2\sigma^2))$ and the Laplacian kernel as $\cK(a, b) = \exp(-\|a-b\|_1/\sigma)$, where $\sigma > 0$ is the bandwidth parameter, $\|\cdot\|_2$ is the Euclidean $\ell_2$-norm, and $\|\cdot\|_1$ is the $\ell_1$-norm.
		
		\section{Conditional Mean Independence Testing}\label{sec2_method}
		In this section, we introduce a novel population measure for evaluating CMI and propose a corresponding sample-based statistic for conducting the CMI test. We then establish theoretical guarantees for the proposed procedure, including size control and power against local alternatives.

		\subsection{Population conditional mean independence measure}\label{sec_cmi}
		
		Recall that the goal is to test the null hypothesis of $H_0: \E[Y|X,Z] = \E[Y|Z]$ a.s.-$\mathbb P_{XZ}$ against the alternative hypothesis of $H_1: \mathbb P\big(\E[Y|X,Z] \neq\E[Y|Z]\big)>0$. To motivate our population measure for evaluating CMI, we begin with the following result, which provides equivalent characterizations of CMI~\citep{daudin1980partial}.
		\begin{proposition}\label{prop_eqve}
			If $\E [\|Y\|_2^2]<\infty$, then the following properties are equivalent to each other:
			\begin{enumerate}[(a)]\vspace{-0.5em}
				\item\label{eqv1} $\E[Y|X,Z] = \E[Y|Z]$ a.s.-$\mathbb P_{XZ}$. 
                \vspace{-0.5em}
				\item\label{eqv3} $\E\big[\big(f(X,Z)-\E[f(X,Z)|Z]\big)\, Y \big] = 0$ for any $f\in L_2(\R^{d_X+d_Z},\bP_{XZ})$.\vspace{-0.5em}
				\item\label{eqv2} $\E\big[\big(f(X,Z)-\E[f(X,Z)|Z]\big) \, \big(Y-\E[Y|Z]\big)  \big] = 0$ for any $f\in L_2(\R^{d_X+d_Z},\mathbb P_{XZ})$. \vspace{-0.5em}
			\end{enumerate}
		\end{proposition}
		
		\begin{remark}
			\textit{  It is straightforward to see that (\ref{eqv1}) $\Rightarrow$ (\ref{eqv3}) $\Rightarrow$ (\ref{eqv2}), and (\ref{eqv2}) implies (\ref{eqv1}) by taking $f(X,Z) = \E[Y^\top c\,|\,X,Z]$ over all $c\in\R^{d_Y}$. The only difference between (\ref{eqv2}) and (\ref{eqv3}) in Proposition \ref{prop_eqve} is that $Y$ is centered at $\E[Y|Z]$ in (\ref{eqv2}). This additional centering is crucial for reducing biases from the estimation of the conditional mean functions; our proposed population CMI measure will be derived from (\ref{eqv2}) by considering all $f$ within a dense subset of $L_2(\R^{d_X+d_Z},\mathbb P_{XZ})$.}
		\end{remark}	
		
		\smallskip
		Let $\mathcal{K}_X:\, \R^{d_X}\times\R^{d_X}$ and $\mathcal{K}_Z:\, \R^{d_Z}\times\R^{d_Z}$ denote two symmetric positive-definite kernel functions that define two reproducing kernel Hilbert spaces (RKHS) $\bH_X$ and $\bH_Z$ over the spaces of $X$ and $Z$, respectively. Additionally, let $\cK_0 = \cK_X\times\cK_Z$ represent the product kernel, with $\bH_0$ being the corresponding RKHS induced by $\cK_0$ over the product space $\R^{d_X}\times \R^{d_Z}$.
	Motivated by part (\ref{eqv2}) of Proposition \ref{prop_eqve}, we define linear operator $\Sigma:\R^{d_Y}\to \bH_{0}$,
       \small\begin{align*}
            \Sigma\, c =&\,  \E\Big\{\Big[\cK_{0}\big((X,Z),\,\cdot\,\big)-\E\big[\cK_{0}\big((X,Z),\,\cdot\,\big)\big|Z\big]\Big]\\
            & \qquad \qquad \big[Y-\E[ Y|Z]\big]^\top c\Big\}, \quad\mbox{for any $c\in\R^{d_Y}$.}
        \end{align*}\normalsize
        From the reproducing property, we see that for any $f\in\bH_0$ and any $c\in\R^{d_Y}$,
        		\small\begin{align}
			\li f,  \Sigma \,c\ri_{\bH_{0}} 
      =&\, \E\Big\{\big[f(X,Z)-\E[f(X,Z)|Z]\big] \big[Y-\E[Y|Z]\big]^\top c  \Big\}.\nonumber
		\end{align}\normalsize
       Under the assumption that the RKHS $\bH_0$ is dense in $L_2(\R^{d_X+d_Z},P_{XZ})$, which holds if $\cK_X$ and $\cK_Z$ are $L_2$- or $c_0$-universal kernels \citep[Theorem 5]{szabo2018characteristic}, such as the Gaussian and Laplacian kernels considered in this paper, the preceding display along with Proposition \ref{prop_eqve} implies that the null hypothesis $H_0$ holds if and only if $\Sigma$ is the zero operator (i.e., $\Sigma c = 0\in\bH_0$ for any $c\in\R^{d_Y}$).
       The following proposition formalizes this intuition and serves as the foundation for our proposed population CMI measure,
       		\small\begin{align}\label{eq_mid1}
			\Gamma^\ast =& \, \E\big[U(X,X')\, V(Y,Y')\,\cK_Z(Z,Z')\big],\\
		\mbox{where} \ \  V(Y,Y')  &\,= \big[Y-g_Y(Z)\big]^\top\big[Y'-g_Y(Z')\big],\ \ \mbox{and} \nonumber\\
			U(X,X') = &\,\cK_X(X, X')  -\li g_X(Z),\cK_X(X',\cdot)\ri_{\bH_X}\nonumber \\
			-\li &\, g_X(Z'),\cK_X(X,\cdot)\ri_{\bH_X}+\li g_X(Z),g_X(Z')\ri_{\bH_X}. \nonumber \\[-4ex]\notag
		\end{align}\normalsize
		Here, $(X',Y',Z')$ is an independent copy of $(X,Y,Z)$, $g_Y$ and $g_X$ are defined as $g_Y(\cdot){=}\E[Y|Z{=}\,\cdot\,]\in\R^{d_Y}$ and $g_X(\cdot){=}\E[\,\cK_X(X,\cdot)\,|\,Z{=}\,\cdot\,]\in\bH_X$, respectively.
\begin{proposition}\label{prop_CMI}
   If Assumption \ref{assump2}(a) holds, then 
    \begin{enumerate}[(a)] \vspace{-0.5em}
\item\label{ppr2} $\Sigma$ is a Hilbert-Schmidt operator, and its Hilbert-Schmidt norm, denoted as $\|\Sigma\|_{\rm HS}$, satisfies $\|\Sigma\|_{\rm HS}^2 = \Gamma^\ast$. \vspace{-0.5em}
\item\label{ppr1} The null $H_0$ holds if and only if $\Gamma^\ast = 0$.
\vspace{-0.2em}
\end{enumerate}
\end{proposition}
Due to the multiplicative form inside the expectation in equation~(\ref{eq_mid1}), when constructing a sample version of $\Gamma^\ast$ as the test statistic, the estimation errors of the two nuisance parameters ($g_X$ and $g_Y$) do not affect the asymptotic properties of the statistic, as long as the product of these errors decays faster than $n^{-1/2}$
. This desirable property is commonly referred to as double robustness~\citep{bang2005doubly,zhang2024doubly}.

Our $\Gamma^\ast$ bears similarity to the maximum mean discrepancy-based conditional independence measure (MMDCI) proposed in \citet{zhang2024doubly}, which was designed for testing the stronger null hypothesis of conditional independence (CI). While CI assumes that the conditional distribution of the response variable $Y$ is independent of an additional covariate $X$ given an existing covariate $Z$, the CMI assumption only requires that the conditional mean function of $Y$ does not depend on $X$ given $Z$. This means that including $X$ in the regression function of $Y$ does not improve its predictive ability in the mean squared error sense. In contrast, rejecting CI does not provide a direct interpretation in terms of predictive ability \citep[Section 1]{lundborg2022projected}.
Furthermore, the MMDCI statistic compares the joint distribution of $(X,Y,Z)$ under the null with its true distribution using maximum mean discrepancy (MMD) \citep{gretton2012kernel}, requiring stronger assumptions to fully characterize CI (see Assumption 1 in \citet{zhang2024doubly}). In contrast, our population CMI measure avoids using MMD to capture the full distributional properties of $(X,Y,Z)$, making it more suitable for CMI testing compared to existing population CMI measures.
Computationally, the test proposed in \citet{zhang2024doubly} requires training two generative neural networks (GNNs) to sample from the conditional distributions of $X$ and $Y$ given $Z$, respectively. In contrast, our proposed test statistic only requires sampling from the conditional distribution of $X$ given $Z$, resulting in lower computational cost and reduced memory usage.

		\subsection{Testing procedure}\label{sec:test}
To construct a sample version of $\Gamma^\ast$, it is necessary to estimate $g_Y$ and $g_X$. To address the curse of dimensionality, we use deep neural networks (DNNs) to estimate $g_Y$. For $g_X$, which is an RKHS-valued function, we estimate $g_X(z)$ for a fixed $z \in \R^{d_Z}$ by sampling $M$ copies $\{X_{z,i}\}_{i=1}^M$ from the conditional distribution of $X$ given $Z=z$. The estimate of $g_X(z)$ is then given by the sample average $M^{-1} \sum_{i=1}^M \cK_X(X_{z,i}, \cdot) \in \cH_X$.
Specifically, using the noise-outsourcing lemma (Theorem 6.10 of \citet{kallenberg2002foundations}; see also Lemma 2.1 of \citet{zhou2022deep}), for any integer $m \geq 1$, there exists a measurable function $\bG$ such that, for any $\eta \sim N(0, I_m)$ independent of $Z$, we have $\bG(\eta, Z) \,|\, Z \sim P_{X|Z}$. Based on this result, we train a generative moment matching network \citep[GMMN,][]{dziugaite2015training, li2015generative} to approximate $\bG$. For any fixed $z \in \R^{d_Z}$, we generate i.i.d. copies ${\eta_i}$ from $N(0, I_m)$, input them into the trained GMMN $\wh \bG$ along with $z$, and treat the outputs $\wh \bG(\eta_i, z)$ as approximately sampled from the conditional distribution of $X$ given $Z=z$.
We refer to the trained GMMN as the (conditional) generator, and detailed training procedures are provided in Appendix. 

In order to eliminate dependence on conditional generator estimation and improve test size accuracy, we follow \citet{shi2021double} and adopt a sample-splitting and cross-fitting framework to train the GMMN and DNN. For easy presentation, we consider two-fold splitting in this paper, although the test can be readily generalized to the multiple-fold splitting setting. Specifically, for positive integers $n$, $M$ and $m \in [M]$, let $\{(X_i,Y_i,Z_i)\}_{i=1}^n$ and $\{\eta_i^m\}_{i=1}^n$ be i.i.d.~copies of $(X,Y,Z)$ and $\eta$ such that $\big\{\{(X_i,Y_i,Z_i)\}_{i=1}^n, \{\eta_i^1\}_{i=1}^n,\dots,\{\eta_i^m\}_{i=1}^n\big\}$ are mutually independent. Let $\wt X_i^{(m)} = \bG(\eta_i^m,Z_i)$ denote the data sampled from $P_{X_i|Z_i}$. We divide $[n]$ into two equal folds $\mathcal{J}^{(1)}$ and $\cJ^{(2)}$ where $\mathcal{J}^{(1)}:=\mathcal{J}^{(-2)} = \{1,2,\dots,\lfloor n/2\rfloor\}$ and $\cJ^{(2)}:=\mathcal{J}^{(-1)}=[n]/\cJ^{(1)}$. For $j\in[2]$, we train a GMMN generator $\wh \bG_{j}$ using data $(X_i,Z_i)$ for $i\in \cJ^{(-j)}$. Similarly, we train a DNN $\wh g_{j}$ using data $(Y_i,Z_i)$ for $i\in \cJ^{(-j)}$ as an estimator of $g_Y$ (four neural networks are trained in total). Let $\wh X_i^{(m)} = \wh \bG_{j}(\eta_i^m,Z_i)$ and $\wh g_Y(Z_i) = \wh g_{j}(Z_i)$ if $i\in\cJ^{(j)}$, then we define the test statistic as
		\footnotesize\begin{align*} 
			\wh T_n {=} \frac{1}{2}\sum_{s=1}^2\Big[\frac{1}{\frac{n}{2}(\frac{n}{2}{-}1)} & {\sum_{\substack{j\neq k\\ j,k\in\cJ^{(s)}}} }	\wh U(X_j,X_k) \wh V(Y_j,Y_k) \cK_Z(Z_j,Z_k) \Big],\\
		\mbox{where}\ \ 
			\wh U(X_j,X_k) {=} &\cK_X(X_j,X_k){-}\frac{1}{M}\sum_{m=1}^{M}\cK_X(X_j,\wh X^{(m)}_k)\nonumber \\
			{-}\frac{1}{M}{\sum_{m=1}^{M}}&\cK_X(X_k,\wh X^{(m)}_j){+}\frac{1}{M}{\sum_{m=1}^{M}}\cK_X(\wh X^{(m)}_j,\wh X^{(m)}_k), \nonumber \\
		\mbox{and}\ \ 	\wh V(Y_j,Y_k) {=} &\big[Y_j{-}\wh g_Y(Z_j)\big]^\top\big[Y_k{-}\wh g_Y(Z_k)\big]. \nonumber
		\end{align*}\normalsize
		As will be shown in Theorem \ref{th_null} below, the limiting null distribution of $n\wh T_n$ depends on the unknown distribution $P_{XYZ}$. Consequently, we cannot directly determine the rejection threshold of $n\wh T_n$ without knowing $P_{XYZ}$. Instead, we employ a wild bootstrap method to approximate the distribution of $n\wh T_n$. Following Section 2.4 of \citet{zhang2018conditional}, for a positive integer $B$ and $b \in [B]$, we generate $\{e_{bi}\}_{i=1}^n$ from a Rademacher distribution, and define the bootstrap version of $\wh T_n$ as 
		\footnotesize\begin{align}
			\wh M_n^b = &\frac{1}{2}\sum_{s=1}^2\Big\{\frac{1}{\frac{n}{2}(\frac{n}{2}{-}1)}\sum_{\substack{j\neq k\\ j,k\in\cJ^{(s)}}} 	\wh U(X_j,X_k) \wh V(Y_j,Y_k) \nonumber \\
			&\qquad\qquad\qquad\qquad\qquad\cdot\cK_Z(Z_j,Z_k) e_{bj}e_{bk}\Big\}. \label{eqbt}
		\end{align}\normalsize
		Since $\{\wh M_n^1,\wh M_n^2,\dots,\wh M_n^B\}$ can be viewed as samples from the distribution of $\wh T_n$ (c.f.~Theorem~\ref{th_boot}), we reject $H_0$ at level $\gamma\in(0,1)$ if $\frac{1}{B}\sum_{b=1}^{B}  \id_{\{n\wh M_n^b >n\wh T_n\}}{<}\gamma$. As a default choice, we set $B=1,000$ throughout our numerical experiments.

		\subsection{Theoretical properties}\label{sec_2_3}
In this part, we evaluate the asymptotic performance of the test as the sample size increases, focusing on its empirical size (Type-I error) control and power against (local) alternatives. Specifically, we aim to determine whether the proposed test satisfies two desirable theoretical properties:
1.~the probability of incorrectly rejecting the null hypothesis converges to the specified significance level $\gamma$ as the sample size grows; 2.~the test maintains the capability to detect (local) alternative hypotheses with a deviation from null that diminishes at the parametric rate of $n^{-1/2}$.   
		
	From our procedure, we define the event ${\rm Rej}_n:=\big\{B^{-1}\sum_{b=1}^{B}  \id_{\{n\wh M_n^b >n\wh T_n\}}{<}\gamma\big\}$ as rejecting $H_0$. To demonstrate that our proposed test has an accurate asymptotic size, it suffices to prove that the probability of rejecting $H_0$, given that $H_0$ is true, converges to the specified level $\gamma$, i.e., $\lim_{n\to \infty} \bP({\rm Rej}_n | H_0) = \gamma$. To achieve this, we first derive the limiting null distribution of $n\wh T_n$ in Theorem \ref{th_null}. Subsequently, we show the consistency of our bootstrap procedure in Theorem \ref{th_boot}, that is, conditional on the sample, the rescaled bootstrap statistic converges to the same limiting null distribution.
    To study the asymptotic power of our proposed test, we define a local alternative hypothesis in equation~(\ref{eq_assump_power}) whose deviation from null scales as $n^{-\alpha}$ for some $\alpha \geq 0$. The asymptotic behavior of $\wh T_n$ and the bootstrap counterpart under different values of $\alpha$ is analyzed in Theorem \ref{th_alter} and Theorem \ref{th_boot}, respectively. Based on these results, we derive the asymptotic power of the test in Theorem \ref{th_boot}.
     Let $T_n$ denote the oracle test statistic, defined similarly to $\widehat T_n$, except that the estimated nuisance parameters $(g_Y, g_X)$ are replaced with their true values; see Appendix 
     for the precise definition.
		
	 The following assumption is needed to derive the asymptotic properties of our statistic. 

        	\begin{assumption}\label{assump2}
			Assume $M\to \infty$ as $n\to \infty$. For $C_0>0$, $\alpha_1,\alpha_2\in(0,\frac{1}{2})$ such that $\alpha_1{+}\alpha_2>\frac{1}{2}$, $D_{i} \in\{X_{i}, \wt X_{i}^{(1)},\wh X_{i}^{(1)}\}$, $E_{i} \in\{Y_{i}, g_Y(Z_{i}),\wh g_Y(Z_{i})\}$ and $i\in \{i_1,i_2\}$ where $i_s\in \cJ^{(s)}$ for each $s\in[2]$, we have:
			\begin{enumerate}[(a)]
				\item  $\E\Big\{\|E_{i}\|_2^2+\cK_X(D_{i},D_{i})\cK_Z(Z_i,Z_i)+\cK_X(D_{i},D_{i})\|E_{i}\|_2^2\cK_Z(Z_i,Z_i)\Big\}{<}C_0$.
				\item\label{assump2_b} \footnotesize $\Big[\E\Big\{\|\E[\cK_X(\cdot,X_{i})|Z_{i}]{-}\E[\cK_X(\cdot,\wh X_{i}^{(1)})|Z_{i}]\|_{\bH_X}^2\big[\sqrt{\cK_Z(Z_{i},Z_{i})}\allowbreak{+}\|E_{i}\|_2^2\cK_Z(Z_{i},Z_{i})\big]\Big\}\Big]^{1/2}{=}O(n^{-\alpha_1})$ and \\$\Big[\E\Big\{\|g_Y(Z_{i}) {-}\wh g_Y(Z_{i})\|_2^2\big[\sqrt{\cK_Z(Z_{i},Z_{i})}{+}\cK_X(D_{i},D_{i})\allowbreak\cK_Z(Z_{i},Z_{i})\big]\Big\}\Big]^{1/2}{=}O(n^{-\alpha_2})$.\normalsize
			\end{enumerate}
		\end{assumption}

		\begin{remark}\label{rmk_2_gmmn}
			\textit{		If $\cK_X$, $\cK_Z$ are bounded kernels and $\|Y\|_2^2$ is bounded (without loss of generality, assume they are bounded by $1$), then Assumption \ref{assump2} reduces to $\E\big[\big\|\E[\cK_X(\cdot, X_{i})\,|\,Z_{i}]-\E\big[\cK_X(\cdot,\widehat X_{i}^{(1)})\,|\,Z_{i}]\big\|_{\cH_X}^2\big] = O(n^{-2\alpha_1})$ and $\E\big[\|g_Y(Z_{i}) {-}\wh g_Y(Z_{i})\|_2^2\big] = O(n^{-2\alpha_2})$. Note that
				\begin{align}
					& \E\big[\big\|\E[\cK_X(\cdot, X_{i})\,|\,Z_{i}]-\E\big[\cK_X(\cdot,\widehat X_{i}^{(1)})\,|\,Z_{i}]\big\|_{\cH_X}^2\big] \nonumber\\
					=&\E\Big[\Big\{ \sup_{f\in\cH_X:\,\|f\|_{\cH_X}\leq 1}\E\big[f(\widehat X_{i}^{(1)})- f( X_{i}) \,\big|\,Z_{i} \big]\Big\}^2 \Big]\nonumber
					\\
					\leq &\E\Big[\Big\{ \sup_{f:\R^{d_X}\to\R:\,\|f\|_{\infty}\leq 1}\E\big[f(\widehat X_{i}^{(1)})- f( X_{i}) \,\big|\,Z_{i} \big]\Big\}^2 \Big] \nonumber \\
					=& 2 \,\E \big[d^2_{\rm TV}(P_{ \widehat X_{i}^{(1)}|Z_{i}},P_{ X_{i}|Z_{i}})\big], \label{eqn:relation_TV}
				\end{align}
				where $\|\cdot\|_{\infty}$ denotes the function supreme norm, and $d_{\rm TV}(\cdot,\cdot)$ denotes the total variation distance. Here, the inequality in the second line is implied by the fact that $\|f\|_{\infty} = \sup_{x\in\R^{d_X}}|f(x)|=\sup_{x\in\R^{d_X}}|\li f,\cK_X(x,\cdot)\ri_{\cH_X}|\leq \|f\|_{\cH_X}\sqrt{\cK_X(x,x)}\leq \|f\|_{\cH_X}$. Therefore, we can also replace the error metrics in Assumption \ref{assump2} by the total variation distance and the mean squared error, i.e.,  $ \E \big[d^2_{\rm TV}(P_{ \widehat X_{i}^{(1)}|Z_{i}},P_{  X_{i}|Z_{i}})\big] =  O(n^{-2\alpha_1})$ and $\E\big[\|g_Y(Z_{i}) {-}\wh g_Y(Z_{i})\|_2^2\big] = O(n^{-2\alpha_2})$, which are common assumptions made in existing works for characterizing qualities of conditional generators and nonparametric regression functions. However, the total variation metric may not be a suitable metric for characterizing the closeness between nearly mutually singular distributions, which happens when data are complex objects such as images or texts exhibiting low-dimensional manifold structures~\citep{tang2023minimax}.}
		\end{remark}
		 The following theorem gives the limiting null distribution of our statistic, the proof of which is provided in Appendix. 
		\begin{theorem}\label{th_null}
			Suppose Assumptions \ref{assump2} holds, then under $H_0$,
			\begin{align}
				\widehat T_n - T_n = &O_p\big(n^{-1}[M^{-1/2}+n^{-\alpha_1}+n^{-\alpha_2}]+n^{-1/2-(\alpha_1+\alpha_2)}\big) \nonumber 
			\end{align}
			and $	n\widehat T_n\stackrel{D}{\to}T^\dagger=  T_1^\dagger+T_2^\dagger$, where $\{T_1^\dagger,T_2^\dagger\}$ are i.i.d.~random variables with $T_1^\dagger = \sum_{s=1}^{\infty}\lambda_s(\chi_s^2-1)$. Here, $\chi_s^2$ are i.i.d.~chi-square random variables with one degree of freedom, and $\lambda_s$'s are eigenvalues of the compact self-adjoint operator on $L_2(\R^{d_X+d_Y+d_Z},P_{XYZ})$ induced by the kernel function $h((X_1,Y_1,Z_1),(X_2,Y_2,Z_2)) = U(X_1,X_2)V(Y_1,Y_2)\cK_Z(Z_1,Z_2)$; that is, there exists orthonormal basis $\{f_i(X_1,Y_1,Z_1)\}_{i=1}^\infty$ of $L_2(\R^{d_X+d_Y+d_Z},P_{XYZ})$ such that 
			\begin{align}
				\E\big[h((X_1,Y_1,Z_1),(x,y,z))f_i(X_1,Y_1,Z_1)\big]= \lambda_if_i(x,y,z).\nonumber
			\end{align}
		\end{theorem}
       From this theorem, we observe that the ``plugged-in'' test statistic $\widehat T_n$ becomes asymptotically equivalent to the oracle statistic $T_n$ as long as $\alpha_1 + \alpha_2 > 1/2$. This demonstrates the property of \emph{double robustness}: the slowly decaying nonparametric estimation errors in $(g_Y, g_X)$ do not compromise the asymptotic accuracy of the test, provided the product of these errors decays faster than $n^{-1/2}$.

	Now let us switch to the asymptotic power of the test under local alternatives. We use the triple $(X^0,Y^0,Z^0)$ to denote $(X,Y,Z)$ under $H_0$ and consider a sequence of triples $(X^0,Y_n^A,Z^0)$ under the alternative hypothesis:
		\begin{align}\label{eq_assump_power}
			H_{1n}: Y_n^A= \E[Y^0|Z^0]+n^{-\alpha}\cG(X^0,Z^0)+\cR_{n}.
		\end{align}
	 Here, the $\R^{d_Y}$-valued function $\cG$ and the random vector $\cR_n$ satisfy $\E[\cG(X^0,Z^0)|Z^0] = 0$ and $\E[\cR_n|X^0,Z^0] = 0$, respectively, so that the exponent $\alpha \geq 0$ determines the decay rate of the deviation from null under $H_{1n}$; for instance, setting $\alpha = 0$ and $\cR_n \equiv \cR$ corresponds to a fixed alternative.
		The following assumption is needed to derive the asymptotic results under $H_{1n}$ when $\alpha>0$.
		\begin{assumption}\label{assump_local}
			For $D\in\{\cK_X(X^0,\cdot),g_X(Z^0)\}$, there exists a random variable $\zeta\in\R^{d_Y}$ such that $\E\Big\{\|D\|_{\bH_X}^2\|\cR_{n}-\zeta\|^2_2\cK_Z(Z^0,Z^0)\Big\}{\to} 0.$
		\end{assumption}
		\begin{remark}
		\textit{	Assumption \ref{assump_local} implies that when $\alpha>0$, the demeaned random vector $\cR_n =Y_n^A- \E[Y_n^A|X^0,Z^0]$ converges to some random vector $\zeta$. Instead of fixing $\cR_n = Y^0- \E[Y^0|Z^0]$ as in nonparametric regression models (see equation (1.1) in \citet{zhu2012dim}), we allow $\cR_n$ to change with $n$ and $\zeta$ can be different from $Y^0- \E[Y^0|Z^0]$.}
		\end{remark}
        \smallskip
		\newcounter{local_alt}
		\begin{theorem}\label{th_alter}
			Suppose Assumptions \ref{assump2} holds, then under $H_{1n}$,
			\begin{enumerate}
				\item If $\alpha=0$, then $\sqrt{n}(\wh T_n-c_0)\stackrel{D}{\to}\frac{1}{\sqrt{2}}\sum_{j=1}^{2}\cG_j^{(0)}$, where \small$c_0 = \E\Big\{U(X_1,X_2)V(Y_1,Y_2)\cK_Z(Z_1,Z_2)\Big\}>0,$\normalsize ~and $\{\cG_1^{(0)},\cG_2^{(0)}\}$ are i.i.d. mean zero normal random variables with variance equal to \small
				$4\var\Big(\E\Big\{U(X_1,X_2)V(Y_1,Y_2)\cK_Z(Z_1,Z_2)\Big|X_2,Y_2,Z_2\Big\}\Big).$\normalsize
				\setcounter{local_alt}{\value{enumi}} 
			\end{enumerate}
			With Assumption \ref{assump_local} further satisfied, we have
			\begin{enumerate}
				\setcounter{enumi}{\value{local_alt}}
				\item If $0{<}\alpha{<} 1/2$, then $n^{2\alpha}\wh T_n\stackrel{p}{\to}c$, where $c = \E \Big\{U(X_1,X_2)\cG(X_1,Z_1)^\top\cG(X_2,Z_2)\cK_Z(Z_1,Z_2)\Big\}{>}0$.
				\item If  $\alpha = 1/2$,  then $n\wh T_n\stackrel{D}{\to}c{+}T^\dagger_A{+}\frac{1}{\sqrt{2}}\sum_{j=1}^{2}\cG_j$, where $T^\dagger_A = \sum_{j=1}^{2}T^\dagger_{Aj}$ and $\{T^\dagger_{A1},T^\dagger_{A2}\}$ are i.i.d. random variables with $T^\dagger_{A1} = \sum_{i=1}^{\infty}\lambda^A_i(\chi_i^2-1)$, $\chi_i^2$ are i.i.d. chi-square random variables with one degree of freedom and $\lambda^A_i$s are the eigenvalues corresponding to the kernel function  
				$h((X_1,\zeta_1,Z_1),(X_2,\zeta_2,Z_2)) = U(X_1,X_2)\zeta_1^\top\zeta_2\cK_Z(Z_1,Z_2)$. Here $\cG_j$ are independent mean zero normal random variables, possibly correlated with $T_{Aj}^\dagger$, with variance equal to
				\small$4\var\Big(\E\Big\{U(X_1,X_2)\big[\cG(X_1,Z_1)^\top\zeta_2 {+}\cG(X_2,Z_2)^\top\zeta_1 \big]\allowbreak\cK_Z(Z_1,Z_2)\Big|\zeta_2,X_2,Z_2\Big\}\Big).$\normalsize
				\item If $\alpha > 1/2$, $n\wh T_n\stackrel{D}{\to}T^\dagger_A$.
			\end{enumerate}
		\end{theorem}
        
Theorem~\ref{th_null} establishes that rejecting $H_0$ if $n\wh T_n$ exceeds the rejection threshold as the $(1-\gamma)$th quantile of the limiting distribution of $T^\dagger$ constitutes a valid test procedure for $H_0$ with an asymptotic size of $\gamma$ for any $\gamma \in (0,1)$. Additionally, items 1 and 2 of Theorem \ref{th_alter} imply that $n\wh T_n$ diverges to infinity under $H_{1n}$ if the deviation from the null hypothesis decays slower than $n^{-1/2}$, ensuring the power of this test procedure to approach one as $n \to \infty$. Together, these properties imply the minimax-optimality of the test procedure based on the test statistic $\wh T_n$, provided the rejection threshold can be computed.
On the other hand, items 3 and 4 of Theorem \ref{th_alter} show that the test has trivial power in detecting alternatives that are too close to the null. This is expected, as even for a parametric linear model, a meaningful CMI test cannot detect local alternatives with deviations from the null decaying faster than the $n^{-1/2}$ parametric rate (e.g., see Section 1.1 of \citet{lundborg2022projected}).

\smallskip
    Finally, we demonstrate that our proposed test in Section~\ref{sec:test} based on bootstrap eliminates the need to compute the rejection threshold, and the resulting test procedure is asymptotically indistinguishable from the optimal test based on $n\wh T_n$.
   For a generic bootstrapped statistic $\wh M_n^{M}$ defined according to equation (\ref{eqbt}) with $\{e_{bi}\}_{i=1}^n$ replaced by independent sample $\{e_i\}_{i=1}^n$ from standard normal distribution,
   we say that $\wh M_n^{M}$ converges in distribution in probability to a random variable $B^\ast$ if, for any subsequence $\wh M_{n_k}^{M_k}$, there exists a further subsequence $\wh M_{n_{k_j}}^{M_{k_j}}$ such that the conditional distribution of $\wh M_{n_{k_j}}^{M_{k_j}}$ given the data $\{X_i,Y_i,Z_i,\eta_i^m\}_{i,m=1}^{\infty}$ converges in distribution to $B^\ast$ almost surely (e.g., see Definition 2.1 of \citet{zhang2018conditional}). We use the notation $\stackrel{D^\ast}{\to}$ to denote convergence in distribution in probability.
		\newcounter{boot_counter}
		\begin{theorem}\label{th_boot}
			Suppose Assumptions \ref{assump2} holds, then we have,
			\begin{enumerate}\vspace{-0.5em}
				\item Under $H_0$, $n \wh M_n^{M}\stackrel{D^\ast}{\to}T^\dagger$.\vspace{-0.5em}
				\item Under $H_{1n}$ with $\alpha=0$, $n \wh M_n^{M}\stackrel{D^\ast}{\to}T_1 = \sum_{j=1}^{2}\wt T_j$,
				where $\{\wt T_1,\wt T_2\}$ are i.i.d random variables with $\wt T_1 = \sum_{i=1}^{\infty}\gamma_i(\chi_i^2-1)$, $\chi_i^2$ are i.i.d chi-square random variables with one degree of freedom and $\gamma_i$s are eigenvalues of 
				$h((X_1,Y_1,Z_1),(X_2,Y_2,Z_2)) = U(X_1,X_2)V(Y_1,Y_2)\cK_Z(Z_1,Z_2)$.
                \vspace{-0.5em}
				\setcounter{boot_counter}{\value{enumi}} 
			\end{enumerate}
			With Assumption \ref{assump_local} further satisfied, we have
			\begin{enumerate}
				\setcounter{enumi}{\value{boot_counter}}\vspace{-0.5em}
				\item Under $H_{1n}$ with $\alpha>0$, $n \wh M_n^{M}\stackrel{D^\ast}{\to}T^\dagger_A$.\vspace{-0.5em}
			\end{enumerate}
		Furthermore, if we let $M^\ast_{nM,\gamma}$ denote the $(1-\gamma)$th quantile of $n\wh M_n^{M}$ conditioning on the data, then the power  (probability of detecting the alternative) of our testing procedure satisfies: if $\alpha{<}1/2$, then $\bP(n\wh T_n\geq M^\ast_{nM,\gamma})\to 1$; if $\alpha=1/2$, then $\bP(n\wh T_n\geq M^\ast_{nM,\gamma})\to \bP(c{+}T^\dagger_A{+}\frac{1}{\sqrt{2}}\sum_{j=1}^{2}\cG_j\geq T_{0,\gamma}^A)$, where $T_{0,\gamma}^A$ denotes the $(1-\gamma)$th quantile of $T^\dagger_A$; if $\alpha>1/2$, then $\bP(n\wh T_n\geq M^\ast_{nM,\gamma})\to \gamma$.
		\end{theorem}
        Since $\bP(n\wh T_n\geq M^\ast_{nM,\gamma}) = \bP({\rm Rej}_n)$,
        item 1 of Theorem \ref{th_boot} demonstrates that the proposed test achieves asymptotically correct size; while items 2 and 3, along with the second part of the theorem, establish that the proposed test is consistent (the probability of rejecting $H_0$ converges to 1) when the alternative lies outside an $n^{-1/2}$-neighborhood of $H_0$ (i.e., when $\alpha<1/2$). Similar to the optimal test based on $n\wh T_n$ discussed after Theorem~\ref{th_alter}, the proposed test will also exhibit trivial power in detecting alternatives that are too close to the null (i.e., when $\alpha > 1/2$).

		\section{Simulation Results}\label{sec3_simu}

		To evaluate the finite-sample performance of our test, we adopt the two examples from \citet{cai2024partial}, where the response variable $Y$ is univariate. Each experiment is repeated $500$ times with sample sizes $n \in \{200, 400, 800\}$. The nominal significance level is set at 5\%.
		
		{\textbf{Example A1:}} \textit{ Consider the linear regression model $Y_i = \beta_Z^\top Z_i + \beta_X^\top X_i + \epsilon_i$ 
		for $i \in [n]$, where $\epsilon_i \stackrel{i.i.d.}{\sim} N(0, 0.5^2)$ are independent of $\{X_i,Z_i\}_{i=1}^n$, $d_X=d_Z=25$ and $(Z_i^\top, X_i^\top ) \stackrel{i.i.d.}{\sim} N(0, \Sigma)$ with the $(i,j)$th element of $\Sigma$ being $\Sigma_{ij} = 0.3^{|i-j|}$ for $i,j\in[50]$. We set the first two components of $\beta_Z$ as one and the rest as zero. For this example, the null $H_0$ corresponds to $\beta_X\equiv 0$. Under the sparse alternative, the first two components of $\beta_X$ are $0.2/\sqrt{2}$ and the rest as zero. Under the dense alternative, every component of $\beta_X$ is fixed at $0.2/\sqrt{25}=0.04$.}
		
		{\textbf{Example A2:}} \textit{ Consider the nonlinear regression model $Y_i = \beta_Z^\top Z_i + \left(\beta_X^\top X_i\right)^2 + \epsilon_i$ for $i \in [n]$, under the same $\beta_Z$ and $\{X_i, Z_i,\epsilon_i\}_{i=1}^n$ as in Example A1. The null $H_0$ also corresponds to $\beta_X\equiv 0$. Under the sparse alternative, we set the first five components of $\beta_X$ to be $10^{-1/2}$ and the rest as zero. Under the dense alternative,  we set the first twelve components of $\beta_X$ as $24^{-1/2}$ and the rest as zero.}
		\begin{table*}[t]
			\caption{Empirical size and size adjusted power for Examples A1 and A2.}\label{tab_sec4_1_Example_A1}
			\centering
			\scriptsize
			\setlength{\tabcolsep}{3.5pt}
			\begin{tabular}{cc|c|cc|cc|cc|cc|cc|c|cc|c|c}
				\toprule
				\midrule
				\multirow{2}{*}{}&\multirow{2}{*}{} & \multirow{2}{*}{$n$} & \multicolumn{2}{c|}{pMIT} &  \multicolumn{2}{c|}{$\text{pMIT}_{e}$} &  \multicolumn{2}{c|}{{$\text{pMIT}_{M}$}} &  \multicolumn{2}{c|}{{$\text{pMIT}_{e_M}$}} &  PCM &  $\text{PCM}_M$ & VIM &  DSP &  $\text{DSP}_M$  &  \multicolumn{1}{c|}{$\wh{T}_n$ } &  \multicolumn{1}{c}{$\wh{T}_n$  }  \\
				&	&	& XGB &	DNN & XGB & DNN & XGB & DNN & XGB & DNN &  &  &  &  &  & Oracle  &     \\  \hline 
				\multirow{9}{*}{Example A1} &	\multirow{3}{*}{$H_0$} 
				& $200$ & $5.6$ & $6.4$ & $5.8$ & $7.0$ & $6.2$ & $8.4$ & $8.0$ & $9.6$ & $2.0$ & $0.0$ & $3.2$ & $0.0$ & $0.0$ & $7.0$ & $8.8$   \\  \cline{3-18}
				&& $400$ & $6.0$ & $4.2$ & $8.0$ & $5.8$ & $11.8$ & $9.6$ & $14.2$ & $13.2$ & $3.4$ & $0.0$ & $4.8$ & $0.0$ & $0.0$ & $6.8$ & $7.0$   \\ \cline{3-18}
				&& $800$ & $7.0$ & $11.2$ & $9.8$ & $12.8$ & $14.2$ & $17.2$ & $20.2$ & $21.0$ & $1.2$ & $0.0$ & $4.8$ & $0.0$ & $0.0$ & $7.2$ & $5.4$   \\ \cline{2-18}
				&	\multirow{3}{*}{$H_1$ sparse} 
				& $200$ & $3.8$ & $9.4$ & $5.6$ & $9.4$ & $3.2$ & $5.6$ & $4.8$ & $7.6$ & $11.0$ & $25.4$ & $12.6$ & $15.8$ & $15.8$ & $88.4$ & $36.0$   \\ \cline{3-18}
				&	& $400$ & $18.6$ & $34.6$ & $30.4$ & $35.8$ & $11.8$ & $55.2$ & $24.8$ & $56.8$ & $27.0$ & $81.8$ & $14.8$ & $14.8$ & $29.8$ & $100$ & $98.6$   \\  \cline{3-18}
				&& $800$ & $91.6$ & $66.0$ & $97.8$ & $66.4$ & $96.0$ & $96.2$ & $99.8$ & $95.8$ & $87.4$ & $99.8$ & $31.6$ & $52.8$ & $91.8$ & $100$ & $100$   \\ \cline{2-18}
				&	\multirow{3}{*}{$H_1$ dense} 
				& $200$ & $6.0$ & $11.6$ & $7.0$ & $11.2$ & $3.0$ & $15.8$ & $4.6$ & $20.0$ & $15.6$ & $30.0$ & $13.2$ & $22.4$ & $43.8$ & $98.4$ & $56.0$   \\ \cline{3-18}
				&	& $400$ & $17.6$ & $63.6$ & $32.2$ & $65.8$ & $11.0$ & $83.6$ & $20.6$ & $86.3$ & $29.2$ & $87.8$ & $17.0$ & $47.2$ & $76.8$ & $100$ & $100$   \\   \cline{3-18}
				&	& $800$ & $86.6$ & $94.8$ & $93.6$ & $95.2$ & $85.6$ & $100$ & $99.0$ & $100$ & $91.2$ & $100$ & $38.6$ & $80.2$ & $98.8$ & $100$ & $100$   \\ \cline{1-18}
				\multirow{9}{*}{Example A2} &	\multirow{3}{*}{$H_0$} 
				& $200$ & $5.6$ & $6.4$ & $5.8$ & $7.0$ & $6.2$ & $8.4$ & $8.0$ & $9.6$ & $2.0$ & $0.0$ & $3.2$ & $0.0$ & $0.0$ & $7.0$ & $8.8$   \\ \cline{3-18}
				&	& $400$ & $6.0$ & $4.2$ & $8.0$ & $5.8$ & $11.8$ & $9.6$ & $14.2$ & $13.2$ & $3.4$ & $0.0$ & $4.8$ & $0.0$ & $0.0$ & $6.8$ & $7.0$   \\ \cline{3-18}
				&	& $800$ & $7.0$ & $11.2$ & $9.8$ & $12.8$ & $14.2$ & $17.2$ & $20.2$ & $21.0$ & $1.2$ & $0.0$ & $4.8$ & $0.0$ & $0.0$ & $7.2$ & $5.4$   \\ \cline{2-18}
				&	\multirow{3}{*}{$H_1$ sparse} 
				& $200$ & $7.4$ & $7.6$ & $13.0$ & $12.2$ & $17.6$ & $10.2$ & $27.0$ & $21.0$ & $10.6$ & $17.2$ & $10.6$ & $46.4$ & $74.6$ & $36.8$ & $19.2$   \\ \cline{3-18}
				&	& $400$ & $27.2$ & $24.8$ & $47.0$ & $29.8$ & $28.4$ & $50.6$ & $57.4$ & $60.4$ & $23.2$ & $82.0$ & $26.0$ & $73.0$ & $90.8$ & $80.0$ & $76.2$  \\    \cline{3-18}
				&	& $800$ & $92.2$ & $42.6$ & $98.6$ & $45.6$ & $97.6$ & $90.0$ & $100$ & $91.6$ & $94.2$ & $100$ & $71.8$ & $99.6$ & $99.8$ & $100$ & $100$   \\ \cline{2-18}
				&	\multirow{3}{*}{$H_1$ dense} 
				& $200$ & $7.6$ & $8.2$ & $12.0$ & $12.8$ & $14.4$ & $9.4$ & $25.0$ & $24.8$ & $6.6$ & $11.2$ & $10.6$ & $59.4$ & $79.4$ & $23.8$ & $17.6$   \\ \cline{3-18}
				&	& $400$ & $14.0$ & $32.0$ & $27.2$ & $36.8$ & $15.0$ & $67.4$ & $33.0$ & $75.8$ & $13.4$ & $43.0$ & $12.6$ & $84.0$ & $96.4$ & $57.4$ & $57.8$  \\    \cline{3-18}
				&	& $800$ & $44.2$ & $60.0$ & $67.6$ & $61.6$ & $39.8$ & $98.4$ & $79.4$ & $99.2$ & $58.0$ & $97.2$ & $31.2$ & $99.6$ & $100$ & $99.8$ & $99.2$  \\ \hline 
			\end{tabular}
		\end{table*}

		For $\wh{T}_n$, we opt to use the Laplacian kernel and the bandwidth parameter for each kernel is selected according to the median heuristic \citep[][Section 8]{gretton2012kernel}. For comparison, we also include the simulation results for the CMI test proposed in \citep[Algorithm 3]{williamson2023general} (denoted as VIM), the single and multiple split statistics proposed in \citep[Algorithm 1 and 1$^{DR}$]{lundborg2022projected} (denoted as PCM and PCM$_M$ respectively), the single and multiple split statistics proposed in \citep[equations (3) and (7)]{dai2022significance} (denoted as DSP and DSP$_M$ respectively), the single/multiple split CMI tests proposed in \citet{cai2024partial} (denoted as pMIT and pMIT$_M$ respectively) as well as their power enhanced versions (denoted as pMIT$_e$ and pMIT$e_M$ respectively). For the four tests proposed in \citet{cai2024partial}, we include the simulation results when the conditional mean functions are learned using eXtreme Gradient Boosting (XGB) and DNN. In addition, we also show the simulation results for an oracle version of our statistic $\wh T_n$, where the true CME and conditional mean function of $Y$ are used instead of their estimators.

		For Example A1, as shown in Table \ref{tab_sec4_1_Example_A1}, VIM and $\wh T_n$ have relatively accurate size under $H_0$, while PCM and DSP (as well as their multiple split versions) are severely undersized. The empirical size for pMIT with XGB estimation method is close to the nominal level, but it is oversized with DNN estimation method. The pMIT tests with multiple split and/or power enhancement all have large size distortions and the size distortion gets larger as $n$ increases. For the size adjusted power, our test $\wh T_n$ outperforms all other tests for all values of $n$ under both sparse and dense alternatives, and VIM has the largest power loss although it also utilizes sample splitting and cross fitting. Note that pMIT$_M$ with DNN estimation method actually has larger power than pMIT$_e$, which is supposed to be the power enhanced version of pMIT.
		
		For Example A2, as shown in Table \ref{tab_sec4_1_Example_A1}, the empirical size results are the same as in Example A1. For the size adjusted power, DSP and DSP$_M$ have the best overall performances (especially when the sample size is small) and our test $\wh T_n$ have similar power performance as DSP and DSP$_M$ when $n=800$. Note that the power performances of the tests proposed in \citet{cai2024partial} depends heavily on the sparsity of the alternative and the estimation method used. The XGB estimation method has better performance under sparse alternative while the DNN estimation method outperforms under the dense alternative.

		\section{Imaging Data Applications}\label{sec5_real}
      We apply our proposed CMI method to identify important facial regions for two computer vision tasks: recognizing facial expressions and predicting age.
		\subsection{Facial expression recognition} \label{sec5_expression}

   In this application, we examine whether covering some region of a facial image will influence the prediction accuracy of facial expression. We use the Facial Expression Recognition 2013 Dataset \citep[FER2013, ][]{goodfellow2013challenges} consisting of $48\times48$ pixel grayscale facial images, each attached with a label from one of seven facial expressions: angry, disgust, fear, happy, sad, surprise, neutral (denoted as Expression 1-7). As in \cite{dai2022significance}, we consider seven cases where different hypothesized regions (HR) are covered: top left corner (TL), nose, right eye, mouth, left eye, eyes, face; see Appendix and Figure \ref{fig_loca} for locations of these HRs.

		After applying the same preprocessing procedure as outlined in Section 6.D of \citet{dai2022significance}, we obtain 11700 image-label pairs which will be the samples used in this application. Let $\{(X_i,Y_i)\}_{i=1}^{11700}$ denote the $48\times48$ pixel facial images and their corresponding labels. Note that for any $i\in[11700]$ and $j\in[7]$, $Y_i\in\R^7$ is an one-hot vector with the $j$th component being one and the rest being zero if Expression $j$ is associated with $X_i$. For each HR, we use $\{Z_i\}_{i=1}^{11700}$ to denote the facial images with the HR covered in black. 
		
\begin{figure}[t]
			\centering
			\includegraphics[scale=0.45]{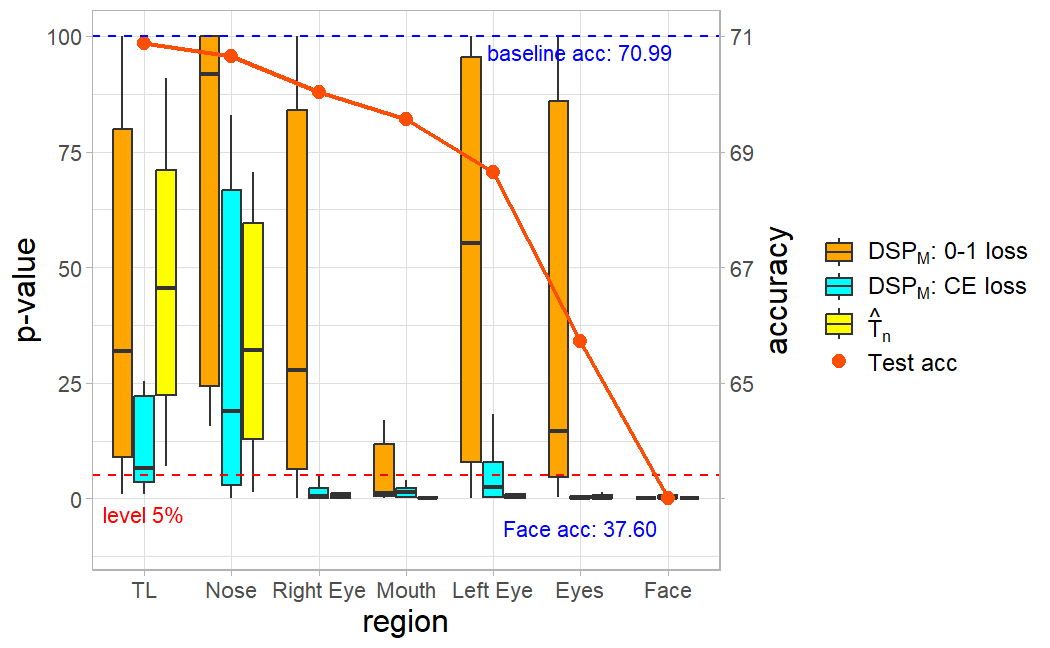}
			\caption{Box plot of the p-values (left y-axis) and the test accuracies (red line, right y-axis) for different HRs. The blue dashed line represents the baseline accuracy. The red dashed line represents the 5\% nominal level. The test accuracy for the face-covered case (face acc: 37.60) is shown at the bottom right corner. }\label{fig_pvalue}
		\end{figure}

		The statistic $\wh T_n$ is evaluated ten times on different subsamples, with sample size $n=2000$, from the 11700 triples $\{(X_i,Y_i,Z_i)\}_{i=1}^{11700}$ and the box plot of the ten p-values are plotted in Figure \ref{fig_pvalue}. As comparison, we also include the p-values of the DSP$_M$ statistics from \cite{dai2022significance} with different loss functions: 0-1 loss and Cross Entropy (CE) loss. To evaluate the testing results, we calculated the test accuracy of a VGG network \citep{khaireddin2021facial} trained/evaluated on the whole sample $\{(Y_i,Z_i)\}_{i=1}^{11700}$, since lower accuracy indicates alternative hypothesis with stronger signal. The resulting test accuracies for different HRs, as well as the accuracy for the VGG network trained/evaluated $\{(X_i,Y_i)\}_{i=1}^{11700}$ (denoted as baseline acc), are also plotted in Figure \ref{fig_pvalue}; see Appendix for the sampling procedure and other implementation details.  
		
		As shown in Figure \ref{fig_pvalue}, the lower quantiles of the p-values from $\wh T_n$ are above the 5\% nominal level when nose or top left corner of the facial image is covered, indicating that these regions are not discriminative to facial expressions. For all the other HRs, the $H_0$ is rejected since all the p-values from $\wh T_n$ are smaller than the nominal level. The test results from $\wh T_n$ is consistent with the test accuracies for different HRs, since the test accuracies when nose or TL is covered are very close to the baseline acc, while for other HRs the test accuracies are noticeably lower than the baseline acc. The test results from DSP$_M$ varies drastically when different loss functions are used. For DSP$_M$ with 0-1 loss, the lower quantiles of the p-values are above (close to) the nominal level for the cases when left eye or both eyes are covered, whereas the test accuracies for these cases are significantly lower than baseline acc, indicating these regions are indeed important in detecting facial expressions. The DSP$_M$ with CE loss has stronger detecting power than DSP$_M$ with 0-1 loss since the p-values for the former is in general smaller than the p-values for the later, which means stronger . However, when TL or nose are covered (which correspond to the cases under $H_0$), the lower quantiles of the p-values from DSP$_M$ with CE loss are smaller than the nominal level, which may result in inflated type-I error. In addition, the median p-values for different HRs from DSP and DSP$_M$ do not decrease monotonously as the test accuracies decrease.

		\begin{figure}[t]
			\centering
			\includegraphics[scale=0.22]{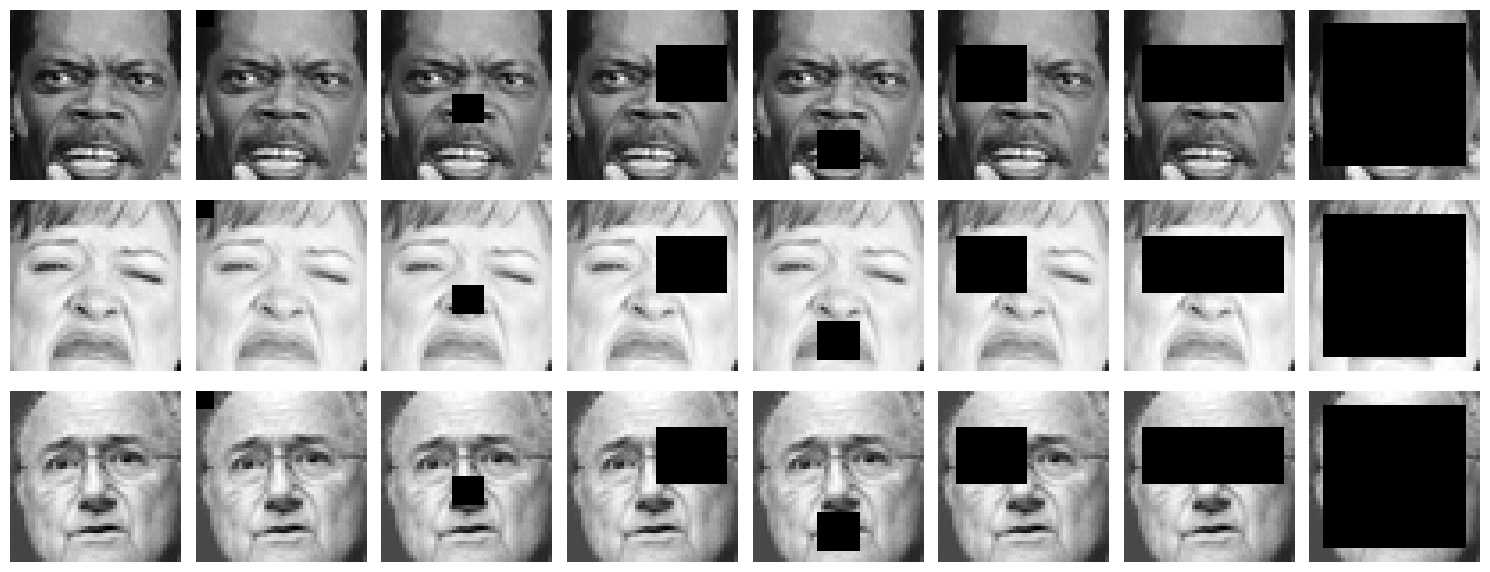}
			\caption{Original facial images in FER2013 (first column) and the covered images with HRs: TL, nose, right eye, mouth, left eye, eyes, face (Columns 2-8).}\label{fig_loca}
		\end{figure}

		\subsection{Facial age estimation} \label{sec5_age}

		In this application, we investigate the impact of covering specific regions of facial images on the accuracy of age prediction using a well-cropped and aligned version of the UTKFace dataset \citep{zhang2017age}, which is available at \url{https://www.kaggle.com/datasets/abhikjha/utk-face-cropped}.  We examine five scenarios where different HRs are covered: the top left corner (TL), nose, mouth, eyes, face; see Appendix and Figure \ref{fig_loca11} for the locations of these HRs.
		
		After converting the images to grayscale and selecting the age labels ranging from 20 to 59 years old, we obtain 16425 image-label pairs, which will be the samples used in this application. Let $\{(X_i,Y_i)\}_{i=1}^{16425}$ denote the $224\times224$ pixel facial images and their corresponding age label $Y_i\in\R$.  For each HR, we use $\{Z_i\}_{i=1}^{16425}$ to denote the facial images with the HR covered in black. 
		
\begin{figure}[t]
			\centering
			\includegraphics[scale=0.45]{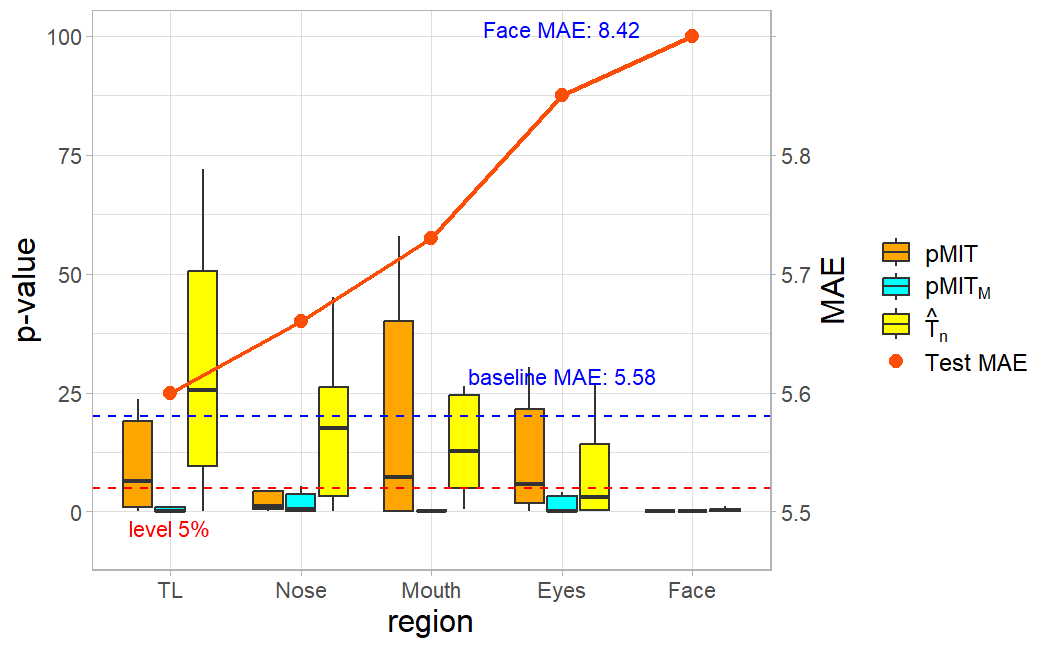}
			\caption{Box plot of the p-values (left y-axis) and the test MAE (red line, right y-axis) for different HRs. The blue dashed line represents the baseline MAE. The red dashed line represents the 5\% nominal level. The test MAE for the face-covered case (face MAE: 8.42) is shown at the bottom right corner. }\label{fig_pvalue11}
		\end{figure}

		The statistic $\wh T_n$ is evaluated ten times on different subsamples, each with a sample size of $n=2000$, from the triples $\{(X_i,Y_i,Z_i)\}_{i=1}^{16425}$. The box plot of the ten p-values is shown in Figure \ref{fig_pvalue11}. For comparison, we also include the p-values of the pMIT and pMIT$_M$ statistics from \citet{cai2024partial}. To evaluate the testing results, we calculated the MAE (mean absolute error) of a EfficientNet B0 network \citep{tan2019efficientnet} trained/evaluated on the whole sample $\{(Y_i,Z_i)\}_{i=1}^{16425}$. The resulting MAE for different HRs, as well as the MAE for the EfficientNet B0 network trained/evaluated using $\{(X_i,Y_i)\}_{i=1}^{16425}$ (denoted as baseline MAE), are also plotted in Figure \ref{fig_pvalue}; see Appendix for the sampling procedure and other implementation details. 
		
		As shown in Figure \ref{fig_pvalue11}, the median of the p-values from $\wh T_n$ for different HRs decrease as the MAEs increase, and they are above the 5\% nominal level when TL, nose, or mouth (three HRs with the smallest MAEs) of the facial image is covered, indicating that these regions are not discriminative for age estimation. For the eyes- and face-covered cases, the null hypothesis $H_0$ is rejected since the median of p-values from $\wh T_n$ are smaller than the nominal level, which makes sense since the test MAE in these two cases are significantly larger than the baseline MAE.

		The p-values from pMIT$_M$ are all lower than the nominal level, even for the TL-covered case where the test MAE is close to the baseline MAE. This is consistent with the simulation result in Section \ref{sec3_simu} where pMIT$_M$ is severely oversized. The p-values from pMIT does not change monotonously with the test MAE. Based on the median p-values, pMIT rejects $H_0$ in the nose-covered case but failed to reject $H_0$ in the month- and eyes-covered cases, even though the later two have larger test MAE than the nose-covered case.

        	
		\begin{figure}[h!]
			\centering
			\includegraphics[scale=0.22]{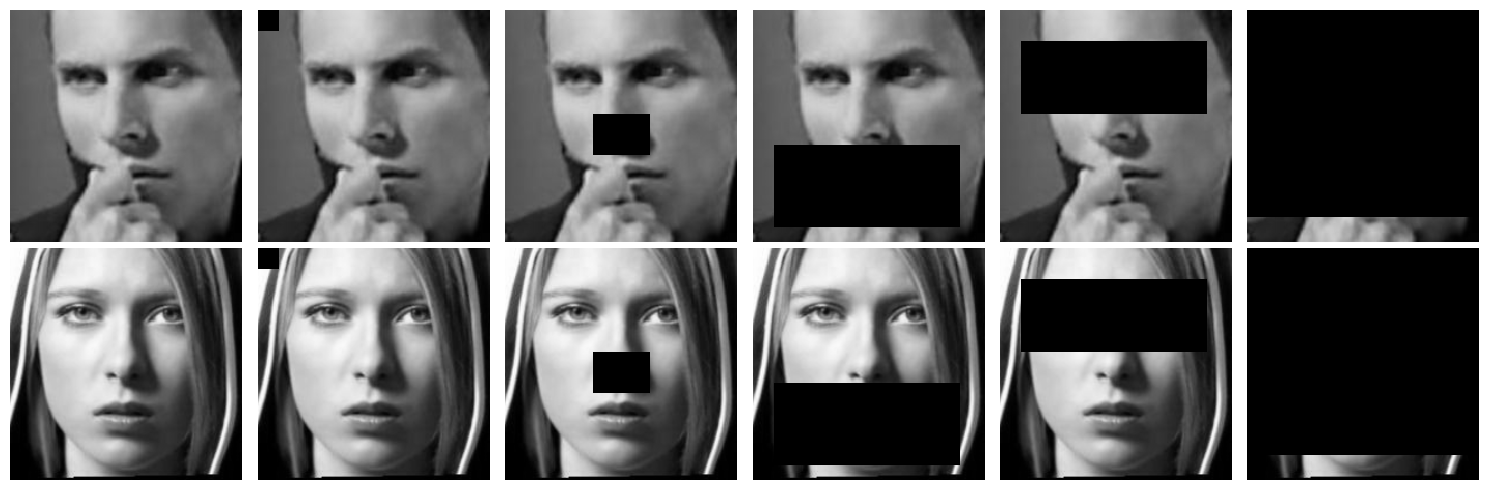}
			\caption{Original facial images in UTKFace (first column) and the covered images with HRs: TL, nose, mouth, eyes, face (Columns 2-6).}\label{fig_loca11}
		\end{figure}

		\section{Conclusion}\label{sec6conclu}
		In this paper, we propose a novel conditional mean independence test that addresses limitations of existing methods. Using RKHS embedding, sample splitting, cross-fitting, and deep generative neural networks, our fully non-parametric test handles multivariate responses, performs well in high-dimensional settings, and maintains power against local alternatives converging at the $n^{-1/2}$ parametric rate.
      Simulations and data applications demonstrate its effectiveness. Some potential future directions include exploring conditional quantile independence testing, developing variable selection methods with false selection rate control, and performing diagnostic checks for high-dimensional regression models.
	       \bibliographystyle{chicago}

	\end{document}